\pdfoutput=1
\documentclass[11pt]{article}

\usepackage{acl}

\usepackage{times}
\usepackage{latexsym}

\usepackage[T1]{fontenc}
\usepackage[utf8]{inputenc}

\usepackage{microtype}

\usepackage{inconsolata}

\usepackage{graphicx}

\usepackage{booktabs}

\title{Probing BERT for German Compound Semantics}

\author{Filip Miletić \hspace{5mm} Aaron Schmid \hspace{5mm} Sabine Schulte im Walde \\
        Institute for Natural Language Processing, University of Stuttgart, Germany \\
        \texttt{\{filip.miletic, aaron.schmid, schulte\}@ims.uni-stuttgart.de}}

\begin{document}
\maketitle
\begin{abstract}
This paper investigates the extent to which pretrained German BERT encodes knowledge of noun compound semantics.
We comprehensively vary combinations of target tokens, layers, and cased vs.\ uncased models, and evaluate them by predicting the compositionality of 868 gold standard compounds.
Looking at representational patterns within the transformer architecture, we observe trends comparable to equivalent prior work on English, with compositionality information most easily recoverable in the early layers.
However, our strongest results clearly lag behind those reported for English, suggesting an inherently more difficult task in German. 
This may be due to the higher productivity of compounding in German than in English and the associated increase in constituent-level ambiguity, including in our target compound set.
\end{abstract}

\section{Introduction}
Noun compounds --  such as \textit{music festival} and \textit{ivory tower} in English; \textit{Obstsaft} ‘fruit juice' and \textit{Sündenbock} ‘scapegoat' (lit.\ ‘sin buck') in German -- comprise a productive class of expressions characterized by variable degrees of compositionality, i.e., relatedness of the individual constituents to the overall meaning of the compound.
The ubiquitousness of noun compounds has motivated a long line of research modeling different aspects of their meanings \citep[][i.a.]{o-seaghdha-2007-annotating, mitchell2008vectorbased, reddy2011empirical, schulteimwalde2016role, cordeiro2019unsupervised},
while more recent work has specifically drawn on their semantically challenging nature to examine the linguistic knowledge encoded in transformer-based language models.

For instance, \citet{garcia2021assessing} question BERT's ability to represent compositionality similarly to humans based on comparisons of compounds in context vs.\ in isolation.
On a more specific level, \citet{garcia2021probing} find a lower quality of BERT representations for non-compositional compounds.
Focusing on semantic relations of noun compounds, \citet{rambelli-etal-2024-large} highlight strong performance variability across large language models as well as difficulties in generalizing to novel compounds, also noted in other related work \citep{li2022systematicity, coil2023chocolate}.

Further studies have attempted to explain these patterns by zooming into the model architecture, but without always reaching a consensus.
As an example, \citet{miletic2023systematic} 
predict the compositionality of open (space-separated) English noun compounds, achieving the best results with embeddings from early transformer layers.
\citet{buijtelaar2023psycholinguistic} similarly predict the semantic transparency of closed (orthographically joined) English noun compounds, but their best results use embeddings from later layers.
Contradictory findings such as these still preclude broader generalizations; 
they are compounded by a limited understanding of cross-lingual trends given a near-exclusive focus on English in prior work \citep{miletic-walde-2024-semantics}.

Moving beyond that focus, we probe BERT via compositionality prediction of 868 German noun-noun compounds \citep{schulte-im-walde-etal-2016-ghost}.
We replicate our setup from \citet{miletic2023systematic} for strict comparability with our prior results on English,
but we introduce a scenario which is more challenging in several key respects:
the closed spelling of German compounds limits constituent-level information in pretraining;
the higher productivity of compounding in German \citep{berg2012determinants} entails more diverse usage contexts and thereby may hinder learning;
and as a result of the higher productivity, the ambiguity of individual constituents may also increase -- including within constituent family sets (i.e., compounds which share one constituent) included in our gold standard data -- and further challenge the models.

We provide a two-fold contribution.
(i)~Comparing all configurations, we broadly find that \textbf{representational patterns generalize cross-lingually}, in particular the relevance of constituent--context comparisons and the recoverability of compositionality information in early transformer layers.
(ii)~Looking at the best configurations, we find that \textbf{BERT's performance on German clearly lags behind English}, which may indicate an inherently more challenging task in German. On a more specific level, this trend may reflect the higher productivity of compounding in German and the related distinctiveness of gold standard information in the two languages.
More generally, our study extends a very limited body of prior work \citep{falk2021automatic, jenkins-etal-2023-split} on German multiword expressions in transformer models.

\section{Data}
\label{sec:data}

\paragraph{Gold standard compounds.}
We rely on the G$_h$oSt-NN dataset of 868 German noun-noun compounds annotated for compositionality, i.e., meaning contributions of the constituents to the overall compound meaning \citep{schulte-im-walde-etal-2016-ghost}.
The targets in the dataset were selected starting from a seed set of 45 compounds balanced for modifier productivity and head ambiguity, and then adding further compounds which contain a modifier or a head already present in the seed set.
By design, the dataset therefore includes constituent family sets, i.e., groups of compounds sharing a constituent.
For example, it contains 15 compounds with the head \textit{Kette} ‘chain', such as \textit{Bergkette} ‘mountain chain', \textit{Hotelkette} ‘hotel chain', and \textit{Halskette} ‘necklace' (lit. ‘neck chain').
Overall, the dataset contains 550 unique modifiers, of which 129 appear more than once; and 279 unique heads, of which 70 appear more than once.

For a given compound--constituent pair, expert annotators were asked to provide a rating from 1 (definitely semantically opaque) to 6 (definitely semantically transparent). 
The averaged final ratings subsume between 5 and 13 individual judgments.
Sample items are shown in Table~\ref{tab:sample-compounds}.

\begin{table}[!h]
    \centering
    \resizebox{\linewidth}{!}{
    \begin{tabular}{|lll|c|c|}
    \hline
    \textbf{Compound} & \textbf{Modif.} & \textbf{Head} & \textbf{M} & \textbf{H} \\
    \hline
    \textit{Erbsensuppe} & \textit{Erbse} & \textit{Suppe} & 5.3 & 5.3 \\
    \textit{\small pea soup} & \textit{\small pea} & \textit{\small soup} && \\
    \hline
    \textit{Kirchspiel} & \textit{Kirche} & \textit{Spiel} & 4.4 & 3.1  \\
    \textit{\small parish} & \textit{\small church} & \textit{\small game} & & \\
    \hline
    \textit{Eifersucht} & \textit{Eifer} & \textit{Sucht} & 2.0 & 2.1  \\
    \textit{\small jealousy} & \textit{\small zeal} & \textit{\small addiction} & & \\
    \hline
    \end{tabular}}
    \caption{Sample compounds and compositionality ratings for the modifier (M) and the head (H).}
    \label{tab:sample-compounds}
\end{table}

\paragraph{Corpus.}
We use the well-established DECOW corpus \citep{schafer2012building, schafer2015processing}
with $\approx11.6$ billion tokens of web-crawled text.
For each compound from the gold standard, we extract all occurrences from the corpus.
In preprocessing, we deterministically split the compound into its constituents by replacing it with the modifier and head provided in the gold standard.
This is done to constrain the output of the pretrained tokenizer used by the BERT models we deploy: it could otherwise split target compounds into subword fragments which are not morphologically motivated \citep[cf.][]{jenkins-etal-2023-split}, which would preclude us from analyzing the model's ability to represent the actual constituents.

\section{Experimental Setup}
\label{sec:exp-setup}

We assess the compositionality information encoded in pretrained BERT via the task of unsupervised compositionality prediction.
We follow the well-established framing of this problem as a ranking task, where a model's ability to represent compound semantics is evaluated by predicting the degrees of compositionality for a set of compounds and correlating those predictions with gold-standard compositionality ratings.
Replicating the experimental setup we introduced in \citet{miletic2023systematic} for English,
we experiment with a wide range of BERT-derived compositionality estimates.
We evaluate each experimental configuration by calculating Spearman's rank correlation coefficient between the predicted degrees of compositionality (based on the cosine score, see below) and the gold-standard compositionality ratings for both modifiers and heads.

\paragraph{BERT models.}
We use the base German BERT model released by DBMDZ%
\footnote{\url{https://huggingface.co/dbmdz/bert-base-german-cased}}
(12 layers, 768 dimensions).
We expand the English setup by comparing the cased and uncased versions of the model given the strong relevance of capitalization for German (nouns are systematically capitalized).
We do not fine-tune the model since our primary aim is to assess the linguistic knowledge it inherently encodes rather than optimize it on the target task.

For a given compound, we feed each corpus example into the model individually.
For each token in the sentence, this yields an embedding corresponding to each layer in the model architecture; we retain all these embeddings.
We then estimate compositionality by comparing pairs of target embeddings in different ways.

\paragraph{Target embeddings.}
We use the following target embeddings:
\texttt{modif}, corresponding to the modifier token;
\texttt{head}, corresponding to the head token;
\texttt{comp}, the average of \texttt{modif} and \texttt{head};
\texttt{cont}, corresponding to the sentence context, i.e., the average of all tokens except for \texttt{modif}, \texttt{head}, \texttt{[CLS]} and \texttt{[SEP]};
\texttt{cls}, corresponding to the \texttt{[CLS]} token which we assume to capture the meaning of the whole sentence.
If the modifier or the head token is split into subwords by BERT's tokenizer, we average over those subwords.

\paragraph{Layers.}
We investigate all available layers, i.e., the input embedding layer and 12 hidden state outputs.
We experiment with all spans of adjacent layers, ranging from a single layer in isolation to the full range of 13 layers, for a total of  91 unique combinations.
When combining embeddings from multiple layers, we average over them.

\paragraph{Compositionality estimates.}
We predict compositionality in two ways.
(i)~Direct estimates correspond to the cosine score for a pair of target embeddings (e.g., \texttt{modif} and \texttt{comp}) from a given layer span.
We test all pairs of target embeddings.
(ii)~Composite estimates use previously proposed composition functions \citep{reddy2011empirical} to combine \texttt{head} and \texttt{modif} predictions obtained with one of the three other target embeddings: \texttt{comp}, \texttt{cont}, and \texttt{cls}.
For example, starting from the cosines for (\texttt{modif}, \texttt{comp}) and (\texttt{head}, \texttt{comp}), we calculate \texttt{ADD} as the sum of the two; \texttt{MULT} as the product of the two; and \texttt{COMB} as the sum of \texttt{ADD} and \texttt{MULT}.

\paragraph{Other settings.}
In order to constrain the experimental space, we only vary the parameters discussed thus far, which we previously found to have a strong effect on model performance in English.
We fix the remaining parameters from our setup in \citet{miletic2023systematic}:
as pooling function, we use averaging over vectors;
we model 100 sentences per compound without controlling for sentence length;
and we use token-level estimates, i.e., we compute compositionality estimates for each sentence individually and then average those estimates to obtain a compound-level value.

\section{Results}
\label{sec:results}

\begin{table}[]
    \centering
    \resizebox{0.9\linewidth}{!}{%
    \begin{tabular}{ccccc}
    \toprule
    & \textbf{Model} & \textbf{Layer} & \textbf{Emb.} & \textbf{$\rho$} \\
    \midrule
    \textbf{Modif.} & uncased & 4--4 & mod, cont & \textbf{0.332} \\
    & uncased & 3--4 & mod, cont & 0.319 \\
    & uncased & 3--5 & mod, cont & 0.317 \\
    & uncased & 4--5 & mod, cont & 0.313 \\
    & uncased & 3--3 & mod, cont & 0.309 \\
    \midrule
    \textbf{Head} & cased & 1--1 & head, cont & \textbf{0.433} \\
    & cased & 1--2 & head, cont & 0.411 \\
    & cased & 0--3 & head, cont & 0.402 \\
    & cased & 1--3 & head, cont & 0.397 \\
    & cased & 0--2 & head, cont & 0.393 \\
    \bottomrule
    \end{tabular}}
    \caption{Best-performing experimental configurations for modifier and head compositionality predictions.}
    \label{tab:best-configs}
\end{table}

\begin{table}[]
    \centering
    \resizebox{\linewidth}{!}{%
    \begin{tabular}{lcc}
    \toprule
    \textbf{Prior approach} & \textbf{Modif.} & \textbf{Head} \\
    \midrule
    \citet{schulteimwalde2016role} & 0.490 & 0.590 \\
    \small{LMI vectors; same data} & & \\
    %\midrule
    \rule{0pt}{3ex}%
    \citet{miletic2023systematic} & 0.553 & 0.645\\
    \small{same method; English data} & & \\
    \bottomrule
    \end{tabular}}
    \caption{Best results reported in prior work.}
    \label{tab:prior-work}
\end{table}

\subsection{Best parameter constellations}
We begin by identifying the best-performing constellations of experimental parameters (Table~\ref{tab:best-configs}).
Our strongest results are weak-to-moderate correlations with gold standard compositionality ratings: $\rho=0.332$ for modifiers and $0.433$ for heads.
But the full set of experimental configurations covers a very broad performance range, reaching negative correlations in the weakest cases ($\rho=-0.159$ for modifiers and $-0.234$ for heads), which confirms that compositionality information is not equally accessible across the BERT architecture.
Furthermore, modifier and head predictions are only weakly correlated with one another ($\rho=0.334$), i.e., the two constituents' respective contributions to the compound meaning are best captured by rather different representational information.

%Mod-head correlation: 0.334
%        mod   head
% mean  0.066  0.153
% min  -0.159 -0.234
% max   0.332  0.433

Looking at prior work (Table~\ref{tab:prior-work}), the higher performance for head than modifier predictions aligns with previously reported trends.
However, our highest results are around $\approx0.2\ \rho$ behind the count-based cooccurrence approach deployed by \citet{schulteimwalde2016role} on the same German dataset.
We also observed a comparable lag of BERT behind simpler vector space approaches for English \citep{miletic2023systematic} with a setup that we replicate here.
But our performance on German also lags behind our prior results for English despite a strictly comparable experiment.
As suggested above, this trend is consistent with an inherently higher difficulty of compositionality prediction in German.
Its more challenging nature could be more specifically due to the higher productivity of compounding in German than in English, which may exacerbate constituent-level ambiguity, including within the G$_h$oSt-NN dataset given its reliance on constituent family sets. 

\begin{table}[!t]
    \centering
    \resizebox{\linewidth}{!}{%
    \begin{tabular}{c|c|ccccc}
    \toprule
    & & \texttt{mod} & \texttt{head} & \texttt{comp} & \texttt{cont} & \texttt{cls} \\
    \midrule
    \textbf{Modif.} & \texttt{mod} &  & \textbf{0.170} & \textbf{0.174} & \cellcolor{Gray!30}\textbf{0.332} & \textbf{0.266} \\
    & \texttt{head} & 0.170 &  & 0.130 & 0.019 & 0.024  \\
    & \texttt{comp} & 0.174 & 0.130 &  & 0.154 & 0.113  \\
    & \texttt{cont} & \cellcolor{Gray!30}\textbf{0.332} & 0.019 & 0.154 &  & 0.123 \\
    & \texttt{cls} & 0.266 & 0.024 & 0.113 & 0.123  \\
    \midrule
    \textbf{Head} & \texttt{mod} &  & 0.327 & 0.202 & 0.178 & 0.084 \\
    & \texttt{head} & \textbf{0.327} &  & 0.290 & \cellcolor{Gray!30}\textbf{0.433} & \textbf{0.246} \\
    & \texttt{comp} & 0.202 & 0.290 &  & 0.318 & 0.149 \\
    & \texttt{cont} & 0.178 & \cellcolor{Gray!30}\textbf{0.433} & \textbf{0.318} &  & 0.096 \\
    & \texttt{cls} & 0.084 & 0.246 & 0.149 & 0.096 \\
    \bottomrule
    \end{tabular}}
    \caption{Best individual results obtained using direct comparisons of pairs of embeddings for modifier predictions (top) and head predictions (bottom). Bold values are best in a column; shaded values are best overall. 
    }
    \label{tab:estimates}
\end{table}

As for the effect of individual experimental parameters, Table~\ref{tab:best-configs} indicates differences between modifier and head predictions regarding the strongest models (uncased vs.\ cased, respectively) and layers (mid-range vs.\ early layers, respectively).
In both cases, the use of embeddings corresponding to the target structure (modifier and head, respectively) in combination with the embedding of the context 
yields the highest results.
Taking a closer look at the interdependency of modifier/head representations and the corresponding predictions, we additionally break down the results across all pairs of target embeddings (Table~\ref{tab:estimates}).
This further confirms the central importance of representational information corresponding to the constituent of interest, closely reflecting prior findings for English \citep{miletic2023systematic}.

\subsection{Cased vs.\ uncased models}

Regarding differences between BERT models, 
modifier predictions are better under experimental configurations using the uncased version (median $\rho=0.060$ vs.\ $0.073$);
by contrast, head predictions benefit from the cased version (median $\rho=0.201$ vs.\ $0.165$).
Looking at the predictions obtained with the cased and uncased model across \textit{all} experimental settings, we find that they are themselves strongly correlated with one another, for modifiers ($\rho=0.768$) as well as heads ($\rho=0.901$).
In other words, the patterns captured by the two model versions are affected by the underlying properties of representational information (embeddings and layers) in a similar -- but not identical -- way.
% corr head: 0.901
% corr mod: 0.768

% case         cased  uncased
% mod  mean    0.059    0.072
%      std     0.075    0.077
%      median  0.060    0.073
% head mean    0.159    0.147
%      std     0.130    0.117
%      median  0.201    0.165

To further understand these interactions, we compute correlations between predictions obtained by the cased vs.\ uncased model in \textit{subsets} of experimental settings.
(i)~We first do this for each of 91 layer combinations.
By keeping the layers fixed, we assess model sensitivity to compositionality estimates.
We find strong mean correlations for modifiers ($0.736\pm0.111$) and heads ($0.915\pm0.055$).
(ii)~We then compute the correlations for each of 19 compositionality estimates.
By keeping the estimates fixed, we assess model sensitivity to layer combinations.
We find moderate mean correlations for modifiers ($0.432\pm0.264$) and heads ($0.494\pm0.213$).
These results indicate that compositionality information captured by the uncased vs.\ cased model is rather similar across compositionality estimates; and rather different across layers.

% keeping layers fixed
% head
% mean    0.915
% std     0.055
% mod
% mean    0.736
% std     0.111

% keeping embs fixed
% head
% mean    0.494
% std     0.213
% mod
% mean    0.432
% std     0.264

\begin{figure}
    \centering
    \includegraphics[width=\linewidth]{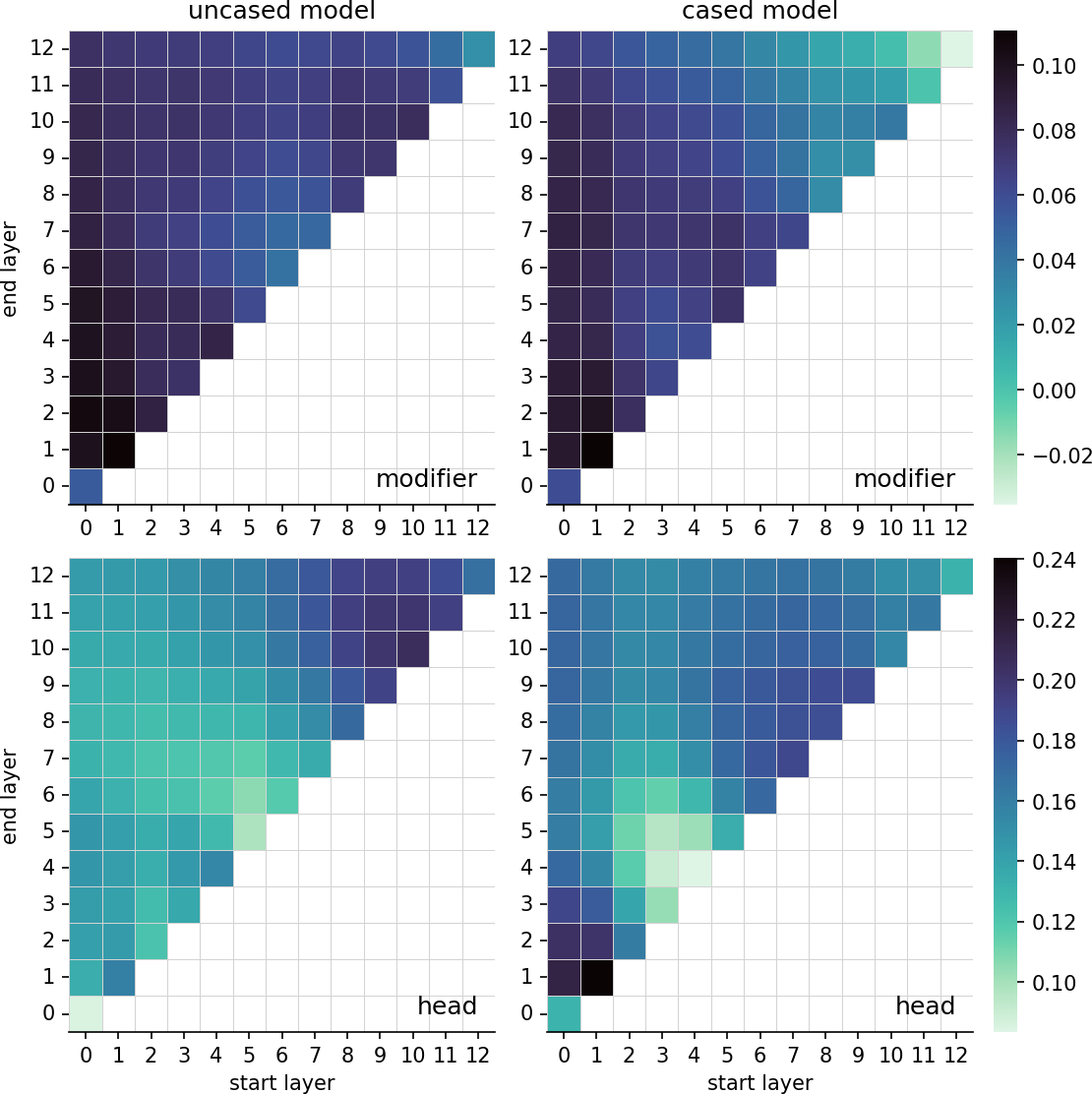}
    \caption{Mean performance across contiguous spans of layers, defined by the start layer (x-axis) and end layer (y-axis). Left: uncased model; right: cased model. Top: modifier predictions; bottom: head predictions.}
    \label{fig:layers}
\end{figure}

\subsection{Layers}
We now examine mean prediction performance for different layer spans to gain further insight into the transformer architecture (Figure~\ref{fig:layers}).
For modifier predictions, the best results are on average obtained in the lower range of layers, with the single highest mean result on layer 1 in isolation ($\rho=0.110$).
Similar performance is obtained by other layer spans -- including very broad ones -- which start from the earliest layers.
By contrast, later layers yield clearly lower results, including layers that are often used in lexical semantic tasks (e.g., layers 9--12, $\rho=0.036$; and layer 12 in isolation, which obtains the single lowest mean $\rho=-0.004$).

Head predictions exhibit comparatively more variance and rather different trends.
Like for modifiers, the single best mean results is on layer 1 in isolation ($\rho=0.199$).
However, the next range of performance
is occupied by spans limited to the very early layers (0--2) and the later layers (7--12), and quite distinctly \textit{not} very broad spans starting from the earliest layers.
The lowest mean result is obtained by input layer 0 in isolation ($\rho=0.108$).

These findings broadly align with good performance of lower layers we reported for English in \citet{miletic2023systematic}, but we find stronger differences between head and modifier predictions.
Some of these interact with the choice of uncased vs.\ cased model; we report the differences between the two models for each layer span in Appendix~\ref{app:layers}, and summarize the trends below.
The uncased model obtains better performance (by $\approx0.05\ \rho$) in the early-to-mid range (layers 3--5, especially for heads) and in the later range (layers 9--12, especially for modifiers).
Put differently, it benefits from stronger contextualization (i.e., processing in more layers), whose disambiguating effect may be relevant given the loss of information inherent in case folding.
The cased model yields gains especially for head predictions, and most clearly in the very early layers (up to $0.08\ \rho$).
We hypothesize that capturing the nominal nature of a constituent -- reflected by capitalization in German, which is preserved by the cased model -- is more important for heads given their dominant role in the morphosyntactic constituency of compounds.

\section{Conclusion}
\label{sec:conclusion}
We investigated the extent to which pretrained German BERT encodes the knowledge of noun compound semantics.
We systematically varied representational information (across target tokens, layers, and cased vs.\ uncased models) to predict the degrees of compositionality of 868 noun-noun compounds.
Our best result ($\rho=0.433$) lags behind equivalent prior work on English -- suggesting a more challenging nature of the task in German -- but we also confirm previously reported patterns of model processing such as the importance of early layers. 
Our insights more generally illustrate the key importance of cross-lingual extensions of probing studies to languages other than English.

\section*{Limitations}
We note several limitations of our work.
(i)~Our study provides a direct cross-lingual comparison with prior results obtained on English, but it is limited to only one other language -- German -- which also belongs the Germanic family and exhibits relatively similar patterns of multiword expression formation.
Typologically more distant languages with stronger structural differences (e.g., Romance languages with a preference for N--Prep--N rather than N--N structures) could provide a further cross-lingual validation of the reported patterns.
(ii)~We only consider noun compounds, but other categories of multiword expressions (e.g., particle verbs) may exhibit different processing patterns in the transformer architecture.
(iii)~We compare the cased and uncased versions of a single German BERT model. Other variations such as German models using different pretraining data or parameter sizes, as well as comparisons with multilingual models, could provide further insights.
(iv)~We compare BERT performance on English and German based on a strictly comparable experimental setup for both languages. However, we use language-specific (and therefore different) gold standard compositionality ratings. While the datasets on which we rely are well-established for each language and define the annotation task in a comparable way, they follow different strategies of selecting target items, which may affect some of the reported trends. For a recent discussion of such effects, see \citet{SchulteImWalde:24}.

\section*{Acknowledgments}
The research presented here was supported by DFG Research Grant SCHU 2580/5-1 (\textit{Computational Models of the Emergence and Diachronic Change of Multi-Word Expression Meanings}).

\clearpage
\bibliography{zotero, new_references}

\begin{thebibliography}{20}
\providecommand{\natexlab}[1]{#1}

\bibitem[{Berg et~al.(2012)Berg, Helmer, Neubauer, and Lohmann}]{berg2012determinants}
Thomas Berg, Sabine Helmer, Marion Neubauer, and Arne Lohmann. 2012.
\newblock \href {https://doi.org/10.1515/ling-2012-0010} {Determinants of the extent of compound use: {{A}} contrastive analysis}.
\newblock \emph{Linguistics}, 50(2).

\bibitem[{Buijtelaar and Pezzelle(2023)}]{buijtelaar2023psycholinguistic}
Lars Buijtelaar and Sandro Pezzelle. 2023.
\newblock \href {https://aclanthology.org/2023.eacl-main.163} {A psycholinguistic analysis of {{BERT}}'s representations of compounds}.
\newblock In \emph{Proceedings of the 17th {{Conference}} of the {{European Chapter}} of the {{Association}} for {{Computational Linguistics}}}, pages 2230--2241, Dubrovnik, Croatia. Association for Computational Linguistics.

\bibitem[{Coil and Shwartz(2023)}]{coil2023chocolate}
Albert Coil and Vered Shwartz. 2023.
\newblock \href {https://doi.org/10.18653/v1/2023.findings-acl.169} {From chocolate bunny to chocolate crocodile: Do language models understand noun compounds?}
\newblock In \emph{Findings of the Association for Computational Linguistics: ACL 2023}, pages 2698--2710, Toronto, Canada. Association for Computational Linguistics.

\bibitem[{Cordeiro et~al.(2019)Cordeiro, Villavicencio, Idiart, and Ramisch}]{cordeiro2019unsupervised}
Silvio Cordeiro, Aline Villavicencio, Marco Idiart, and Carlos Ramisch. 2019.
\newblock \href {https://doi.org/10.1162/coli_a_00341} {Unsupervised compositionality prediction of nominal compounds}.
\newblock \emph{Computational Linguistics}, 45(1):1--57.

\bibitem[{Falk et~al.(2021)Falk, Strakatova, Huber, and Hinrichs}]{falk2021automatic}
Neele Falk, Yana Strakatova, Eva Huber, and Erhard Hinrichs. 2021.
\newblock \href {https://aclanthology.org/2021.iwcs-1.23} {Automatic classification of attributes in {{German}} adjective-noun phrases}.
\newblock In \emph{Proceedings of the 14th {{International Conference}} on {{Computational Semantics}} ({{IWCS}})}, pages 239--249, Groningen, The Netherlands (online). Association for Computational Linguistics.

\bibitem[{Garcia et~al.(2021{\natexlab{a}})Garcia, Kramer~Vieira, Scarton, Idiart, and Villavicencio}]{garcia2021assessing}
Marcos Garcia, Tiago Kramer~Vieira, Carolina Scarton, Marco Idiart, and Aline Villavicencio. 2021{\natexlab{a}}.
\newblock \href {https://doi.org/10.18653/v1/2021.acl-long.212} {Assessing the representations of idiomaticity in vector models with a noun compound dataset labeled at type and token levels}.
\newblock In \emph{Proceedings of the 59th {{Annual Meeting}} of the {{Association}} for {{Computational Linguistics}} and the 11th {{International Joint Conference}} on {{Natural Language Processing}} ({{Volume}} 1: {{Long Papers}})}, pages 2730--2741, Online. Association for Computational Linguistics.

\bibitem[{Garcia et~al.(2021{\natexlab{b}})Garcia, Kramer~Vieira, Scarton, Idiart, and Villavicencio}]{garcia2021probing}
Marcos Garcia, Tiago Kramer~Vieira, Carolina Scarton, Marco Idiart, and Aline Villavicencio. 2021{\natexlab{b}}.
\newblock \href {https://doi.org/10.18653/v1/2021.eacl-main.310} {Probing for idiomaticity in vector space models}.
\newblock In \emph{Proceedings of the 16th {{Conference}} of the {{European Chapter}} of the {{Association}} for {{Computational Linguistics}}: {{Main Volume}}}, pages 3551--3564, Online. Association for Computational Linguistics.

\bibitem[{Jenkins et~al.(2023)Jenkins, Mileti{\'c}, and Schulte~im Walde}]{jenkins-etal-2023-split}
Chris Jenkins, Filip Mileti{\'c}, and Sabine Schulte~im Walde. 2023.
\newblock \href {https://doi.org/10.18653/v1/2023.emnlp-main.1002} {To split or not to split: Composing compounds in contextual vector spaces}.
\newblock In \emph{Proceedings of the 2023 Conference on Empirical Methods in Natural Language Processing}, pages 16131--16136, Singapore. Association for Computational Linguistics.

\bibitem[{Li et~al.(2022)Li, Carlson, and Potts}]{li2022systematicity}
Siyan Li, Riley Carlson, and Christopher Potts. 2022.
\newblock \href {https://aclanthology.org/2022.findings-emnlp.50} {Systematicity in {{GPT-3}}'s interpretation of novel {{English}} noun compounds}.
\newblock In \emph{Findings of the Association for Computational Linguistics: {{EMNLP}} 2022}, pages 717--728, Abu Dhabi, United Arab Emirates. Association for Computational Linguistics.

\bibitem[{Mileti{\'c} and {Schulte im Walde}(2023)}]{miletic2023systematic}
Filip Mileti{\'c} and Sabine {Schulte im Walde}. 2023.
\newblock \href {https://aclanthology.org/2023.eacl-main.110} {A systematic search for compound semantics in pretrained {{BERT}} architectures}.
\newblock In \emph{Proceedings of the 17th {{Conference}} of the {{European Chapter}} of the {{Association}} for {{Computational Linguistics}}}, pages 1499--1512, Dubrovnik, Croatia. Association for Computational Linguistics.

\bibitem[{Mileti{\'c} and Schulte~im Walde(2024)}]{miletic-walde-2024-semantics}
Filip Mileti{\'c} and Sabine Schulte~im Walde. 2024.
\newblock \href {https://doi.org/10.1162/tacl_a_00657} {Semantics of multiword expressions in transformer-based models: A survey}.
\newblock \emph{Transactions of the Association for Computational Linguistics}, 12:593--612.

\bibitem[{Mitchell and Lapata(2008)}]{mitchell2008vectorbased}
Jeff Mitchell and Mirella Lapata. 2008.
\newblock \href {https://aclanthology.org/P08-1028} {Vector-based models of semantic composition}.
\newblock In \emph{Proceedings of {{ACL-08}}: {{HLT}}}, pages 236--244, Columbus, Ohio. Association for Computational Linguistics.

\bibitem[{{\'O}~S{\'e}aghdha(2007)}]{o-seaghdha-2007-annotating}
Diarmuid {\'O}~S{\'e}aghdha. 2007.
\newblock \href {https://aclanthology.org/P07-3013} {Annotating and learning compound noun semantics}.
\newblock In \emph{Proceedings of the {ACL} 2007 Student Research Workshop}, pages 73--78, Prague, Czech Republic. Association for Computational Linguistics.

\bibitem[{Rambelli et~al.(2024)Rambelli, Chersoni, Collacciani, and Bolognesi}]{rambelli-etal-2024-large}
Giulia Rambelli, Emmanuele Chersoni, Claudia Collacciani, and Marianna Bolognesi. 2024.
\newblock \href {https://doi.org/10.18653/v1/2024.acl-long.637} {Can large language models interpret noun-noun compounds? a linguistically-motivated study on lexicalized and novel compounds}.
\newblock In \emph{Proceedings of the 62nd Annual Meeting of the Association for Computational Linguistics (Volume 1: Long Papers)}, pages 11823--11835, Bangkok, Thailand. Association for Computational Linguistics.

\bibitem[{Reddy et~al.(2011)Reddy, McCarthy, and Manandhar}]{reddy2011empirical}
Siva Reddy, Diana McCarthy, and Suresh Manandhar. 2011.
\newblock \href {https://aclanthology.org/I11-1024} {An empirical study on compositionality in compound nouns}.
\newblock In \emph{Proceedings of 5th International Joint Conference on Natural Language Processing}, pages 210--218, Chiang Mai, Thailand. Asian Federation of Natural Language Processing.

\bibitem[{Sch{\"a}fer(2015)}]{schafer2015processing}
Roland Sch{\"a}fer. 2015.
\newblock \href {http://rolandschaefer.net/?p=749} {Processing and querying large web corpora with the {{COW14}} architecture}.
\newblock In \emph{Proceedings of {{Challenges}} in the {{Management}} of {{Large Corpora}} 3 ({{CMLC-3}})}, Lancaster.

\bibitem[{Sch{\"a}fer and Bildhauer(2012)}]{schafer2012building}
Roland Sch{\"a}fer and Felix Bildhauer. 2012.
\newblock \href {https://aclanthology.org/L12-1497/} {Building large corpora from the web using a new efficient tool chain}.
\newblock In \emph{Proceedings of the {{Eighth International Conference}} on {{Language Resources}} and {{Evaluation}} ({{LREC}} 2012)}, pages 486--493, Istanbul, Turkey. European Language Resources Association.

\bibitem[{{Schulte im Walde}(2024)}]{SchulteImWalde:24}
Sabine {Schulte im Walde}. 2024.
\newblock {Collecting and investigating features of compositionality ratings}.
\newblock In Voula Giouli and Verginica Barbu~Mititelu, editors, \emph{Multiword Expressions in Lexical Resources. Linguistic, Lexicographic and Computational Perspectives}, Phraseology and Multiword Expressions. {Language Science Press}, {Berlin}.

\bibitem[{{Schulte im Walde} et~al.(2016){Schulte im Walde}, H{\"a}tty, and Bott}]{schulteimwalde2016role}
Sabine {Schulte im Walde}, Anna H{\"a}tty, and Stefan Bott. 2016.
\newblock \href {https://doi.org/10.18653/v1/S16-2020} {The role of modifier and head properties in predicting the compositionality of {{English}} and {{German}} noun-noun compounds: {{A}} vector-space perspective}.
\newblock In \emph{Proceedings of the {{Fifth Joint Conference}} on {{Lexical}} and {{Computational Semantics}}}, pages 148--158, Berlin, Germany. Association for Computational Linguistics.

\bibitem[{Schulte~im Walde et~al.(2016)Schulte~im Walde, H{\"a}tty, Bott, and Khvtisavrishvili}]{schulte-im-walde-etal-2016-ghost}
Sabine Schulte~im Walde, Anna H{\"a}tty, Stefan Bott, and Nana Khvtisavrishvili. 2016.
\newblock \href {https://aclanthology.org/L16-1362} {{G}ho{S}t-{NN}: A representative gold standard of {G}erman noun-noun compounds}.
\newblock In \emph{Proceedings of the Tenth International Conference on Language Resources and Evaluation ({LREC}'16)}, pages 2285--2292, Portoro{\v{z}}, Slovenia. European Language Resources Association (ELRA).

\end{thebibliography}

\appendix

\clearpage
\onecolumn
\section{Layer performance}
\label{app:layers}
\begin{figure*}[!h]
    \centering
    \includegraphics[width=0.6\linewidth]{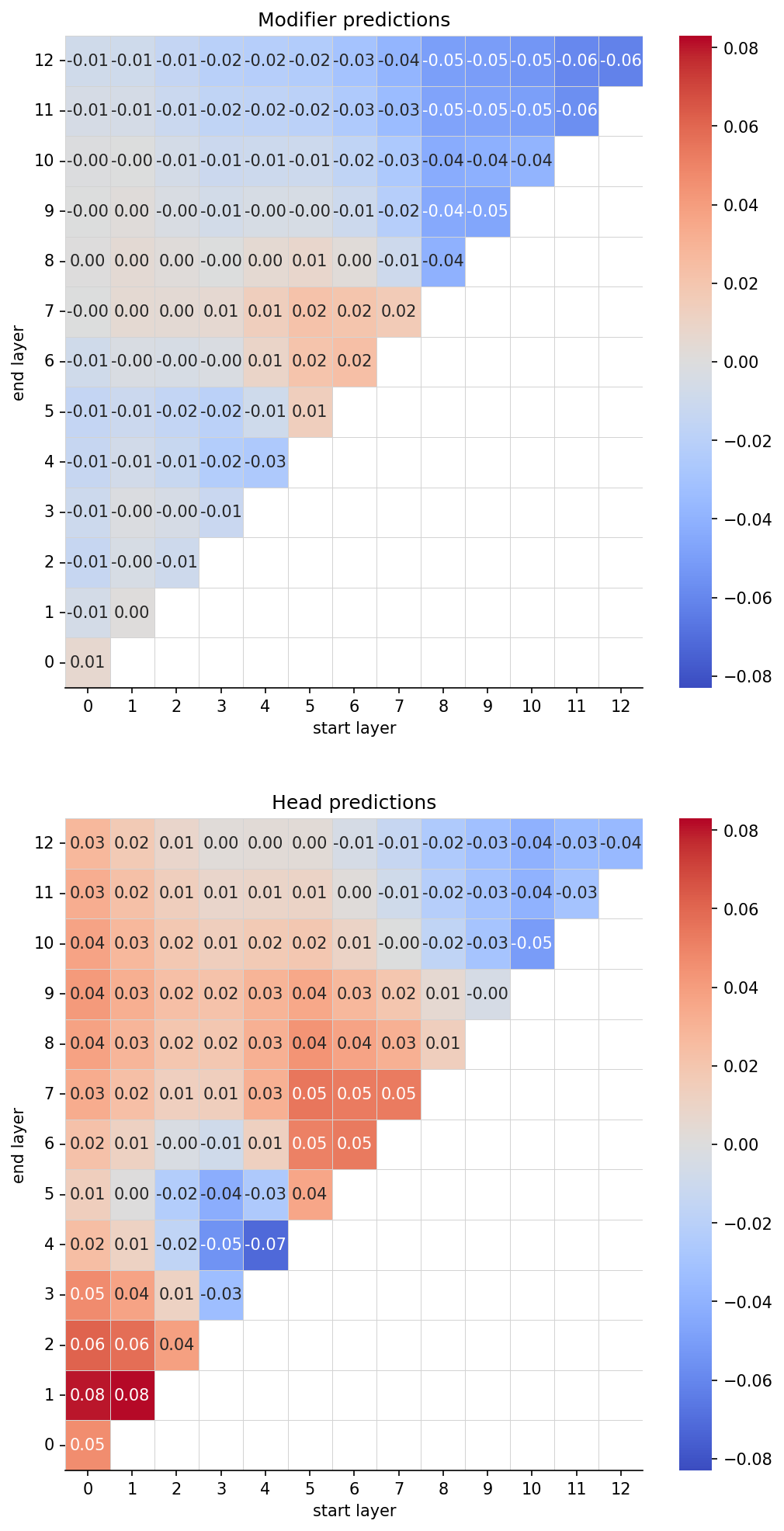}
    \caption{Layer-wise difference in cased vs. uncased model performance. \textbf{Positive values:} better performance of the cased model. \textbf{Negative values:} better performance of the uncased model.}
    \label{fig:layers-deltacased}
\end{figure*}
\end{document}